\documentclass[journal]{IEEEtai}
\usepackage[colorlinks,urlcolor=blue,linkcolor=blue,citecolor=blue]{hyperref}
\usepackage{color,array}
\usepackage{graphicx}
\usepackage{subfigure}
\usepackage{booktabs}

\usepackage{amsmath,amssymb,amsfonts}
\usepackage{algorithmic}
\usepackage{fix-cm}
\usepackage{algorithm,algorithmic}
\usepackage{textcomp}

\begin{document}
\title{Estimating Quality in Therapeutic Conversations: A Multi-Dimensional Natural Language Processing Framework}

\author{Alice Rueda*, \IEEEmembership{Member, IEEE}, Argyrios Perivolaris*, Niloy Roy*, Dylan Weston*, Sarmed Shaya*, Zachary Cote*, Martin Ivanov, Bazen G. Teferra, Yuqi Wu, Sirisha Rambhatla, Divya Sharma, Andrew Greenshaw, Rakesh Jetly, Yanbo Zhang, Bo Cao, Reza Samavi, Sridhar Krishnan\textsuperscript{$\dagger$}, \IEEEmembership{Member, IEEE}, and Venkat Bhat\textsuperscript{$\dagger$}
\thanks{ This paper began as the capstone design project by Dr. Alice Rueda and her students, Niloy Roy, Dylan Weston, Zack Cote, and Sarmed Shaya.}
\thanks{*Co-first authors, \textsuperscript{$\dagger$}Senior co-authors.}
\thanks{This manuscript was submitted to the IEEE Journal of Biomedical and Health Informatics on April 19, 2025. This manuscript did not receive any external funding. AR and BGT are funded by the Canadian Institutes of Health Research (CIHR). VB is supported by an Academic Scholar Award from the University of Toronto Department of Psychiatry and has received research funding from the CIHR, Brain \& Behavior Foundation, Ontario Ministry of Health Innovation Funds, Royal College of Physicians and Surgeons of Canada, Department of National Defence (Government of Canada), New Frontiers in Research Fund, Associated Medical Services Inc. Healthcare, American Foundation for Suicide Prevention, Roche Canada, Novartis, and Eisai.}
\thanks{A. Rueda, A. Perivolaris, M. Ivanov, B.G. Teferra, and V. Bhat are with St. Michael's Hospital, Unity Health Toronto, 30 Bond St., Toronto, Canada M5B 1W8 (e-mail: alice.rueda, argryios.perivolaris, martin.ivanov, bazengashaw.teferra, and venkat.bhat@unityhealth.to). }
\thanks{N. Roy, D. Weston, S. Shaya, Z.Cote, R. Zamavi, and S. Krishnan are with Toronto Metropolitan University, 350 Victoria St., Toronto, Canada M5B 2K3 (e-mail: krishnan@torontomu.ca).}
\thanks{S. Rambhatla is with the University of Waterloo, 200 University Ave W, Waterloo, ON, Canada N2L 3G1 (email: sirisha.rambhatla@uwaterloo.ca)}
\thanks{D.Sharma is with York University, 4700 Keele St, North York, ON, Canada M3J 1P3 (email:divya03@yorku.ca)}
\thanks{A. Greenshaw, Y. Zhang, B. Cao, and Y. Wu are with the University of Alberta, 116 St \& 85 Ave, Edmonton, AB T6G 2R3 (email: andy.greenshaw, yuqi14, yanbo9, cloud.cao@ualberta.ca)}
\thanks{R. Jetly is with the Institute of Mental Health Research, University of Ottawa, 1745 Alta Vista Drive, Ottawa, ON, Canada K1A 0K6 (email: rakesh@drjetly.com )}
}

\maketitle
\begin{abstract}
Engagement between client and therapist is a critical determinant of therapeutic success. We propose a multi-dimensional natural language processing (NLP) framework that objectively classifies engagement quality in counseling sessions based on textual transcripts. Using 253 motivational interviewing transcripts (150 high-quality, 103 low-quality), we extracted 42 features across four domains: conversational dynamics, semantic similarity as topic alignment, sentiment classification, and question detection. Classifiers, including Random Forest (RF), CatBoost, and Support Vector Machines (SVM), were hyperparamter tuned and trained using a stratified 5-fold cross-validation and evaluated on a holdout test set. On balanced (non-augmented) data, RF achieved the highest classification accuracy (76.7\%), and SVM achieved the highest AUC (85.4\%). After SMOTE-Tomek augmentation, performance improved significantly: RF achieved up to 88.9\% accuracy, 90.0\% F1--score, and 94.6\% AUC, while SVM reached 81.1\% accuracy, 83.1\% F1--score, and 93.6\% AUC. The augmented data results reflect the potential of the framework in future larger-scale applications. Feature contribution revealed conversational dynamics and semantic similarity between clients and therapists were among the top contributors, led by words uttered by the client (mean and standard deviation). The framework was robust across the original and augmented datasets and demonstrated consistent improvements in F1 scores and recall. While currently text-based, the framework supports future multimodal extensions (e.g., vocal tone, facial affect) for more holistic assessments. This work introduces a scalable, data-driven method for evaluating engagement quality of the therapy session, offering clinicians real-time feedback to enhance the quality of both virtual and in-person therapeutic interactions.

\end{abstract}

\begin{IEEEkeywords}
NLP features, motivational interviewing, engagement, therapy session, conversation dynamics, semantic similarity, sentiment analysis, question detection
\end{IEEEkeywords}



\section{Introduction}
\label{Introduction}

The COVID-19 pandemic underscored the potential of virtual care, with virtual psychotherapy sessions rising dramatically from 38\% pre-pandemic to 98\% during the pandemic~\cite{Wosik2020}~\cite{Sampaio2021}. However, the shift to virtual platforms highlights a critical challenge: ensuring meaningful engagement between clients and therapists. This shift brought attention to the critical challenge of fostering meaningful engagement between clients and therapists in virtual settings. Engagement, a cornerstone of effective psychotherapy, is intrinsically tied to therapeutic outcomes~\cite{Cadel2021, Duong2021}. Yet, current practices for measuring engagement primarily rely on subjective, retrospective self-reports that are prone to response biases~\cite{Tryon2008} and ceiling effects~\cite{Paap2017}. These limitations necessitate the development of objective, real-time methods to assess engagement during therapy sessions. Our framework will provide engagement level estimation such that the therapists can act accordingly to improve the quality and the outcome of virtual therapy sessions. Currently, there is no consensus on an objective engagement measurement for virtual therapeutic consultation~\cite{Bijkerk2023}. 

A consistent finding has shown that the quality of the alliance~\cite{wampold2023} and engagement~\cite{Bourke2021} are related to subsequent therapeutic outcomes. Therapeutic alliance is an important factor in fostering therapeutic engagement~\cite{Tetley2011}. Therapeutic engagement is critical for the success of any psychotherapy intervention~\cite{Oetzel2003}. When a patient connects and trusts the therapist, the outcomes have proven more positive as patients often attribute positive outcomes to the personal attributes of their therapists~\cite{Horvath2001}. Engagement is useful to our understanding of successful communicative interactions and is a necessary element that denotes a level of commitment to and interest in the therapeutic process ~\cite{Simmons-Mackie2009}. In this work, we focus on an engagement classification based on a collection of metrics from natural language processing (NLP) and the analysis of text transcripts from the motivational interviewing counseling dataset. 

Research on engagement quantification through NLP is well documented in education~\cite{lee2018, Liu2022, toti2021detection, felix2022exploring, slater2017using}, business~\cite{garg2021pulse, morio2023nlp}, social robotics~\cite{Pattar2019, anzalone2015evaluating, Mazzei2022, Hameed2016}, and digital health ~\cite{funk2020, Malins2022}. A study by Lee \textit{et al.}~\cite{lee2018} utilized NLP tools to analyze student blogs and survey responses, employing the Linguistic Inquiry and Word Count (LIWC) tool to measure dimensions like Clout, Authenticity, and Emotional Tone, which indicated student engagement. Liu \textit{et al.}~\cite{Liu2022} explored the relationship between emotional and cognitive engagement in MOOC discussions, finding that positive and confused emotions strongly predicted learning outcomes. Their study used a deep learning model, BERT-CNN, on a dataset of 8,867 learners and 60,624 posts, highlighting the role of sentiment in engagement~\cite{Liu2022}. 

In business, Ma \textit{et al.}~\cite{ma2022investigating} examined brand engagement on social media using NLP. They discovered that interactivity and positive sentiment in Weibo posts drove engagement, although multimedia usage like videos was less impactful~\cite{ma2022investigating}. Social robotics research, such as Pattar \textit{et al.} introduced multimodal systems integrating Autoencoders and Dialogflow for engagement classification in Human-Robot Interaction~\cite{Pattar2019}. In digital health, Funk \textit{et al.} applied NLP to predict user engagement and therapeutic outcomes in interventions, achieving moderate accuracy in predicting behaviors such as binge eating ~\cite{funk2020}. Similarly, Malins \textit{et al.} used machine learning to rate patient activation levels in therapy sessions, with models achieving accuracies of 73\% and 81\%. Together, these studies underline the versatility and potential of NLP frameworks in quantifying engagement across varied applications~\cite{Malins2022}.

Much work on engagement quantification systems exists, however, to the best of our knowledge, no framework for engagement scoring exists in a therapeutic consultation setting. In this study, we address these gaps by introducing a multi-dimensional NLP framework that quantifies engagement in therapeutic conversations. Key contributions include (1) feature extraction across conversational dynamics, semantic similarity, sentiment, and question detection, and (2) interpretability of engagement metrics through SHAP (SHapley Additive exPlanations)~\cite{NIPS2017_7062} value analysis to illustrate top contributing features. 

The objective of this research paper is to determine the engagement quality of the conversation between the client and therapist to reflect the efficacy of counseling sessions. To address this need, a custom framework has been developed, combining machine learning and NLP techniques that can analyze and understand human language (in the form of chat transcripts) to determine the conversation quality during these sessions.

\section{Method}
\begin{figure*}[htb]
\centering
\includegraphics[width=0.7\linewidth, trim={15 10 15 10}, clip]{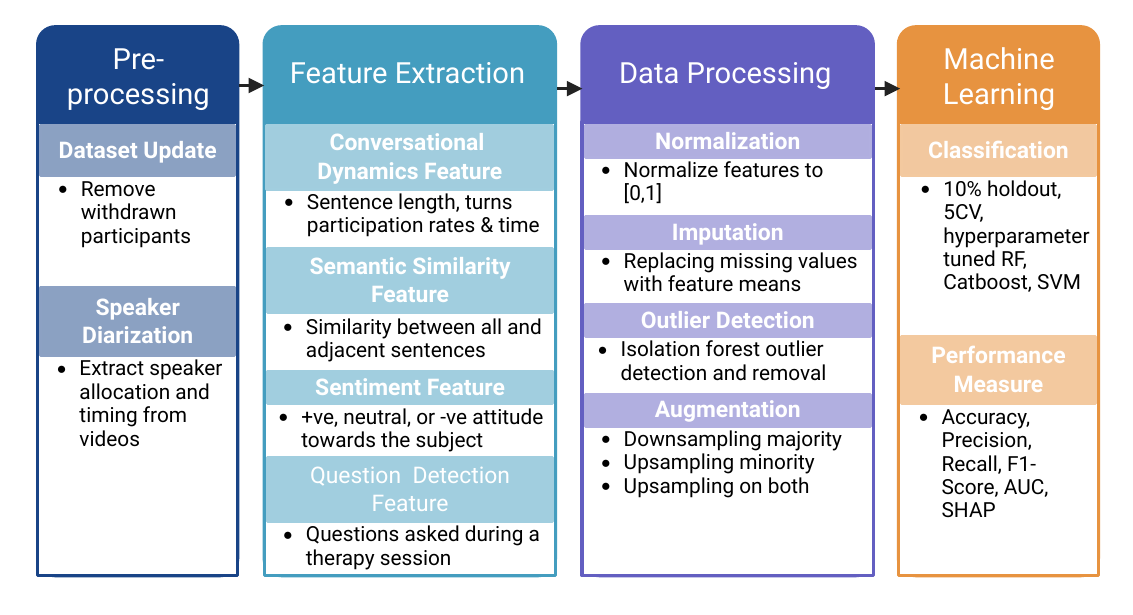}
\caption{Overview of the conversation quality assessment framework. Diarization timing and terms were extracted from the available videos. The system processes transcripts through four feature domains (conversational dynamics, semantic similarity, sentiment, and question detection), followed by data processing steps, before classification using Random Forest, CatBoost and SVM. Finishing with performance measure.}
\label{fig:processing-flow}
\end{figure*}

This study used the motivational interviewing counseling dataset~\cite{perez-rosas-etal-2019-makes} collected from various counseling conversations. The analytical process included feature extraction, data processing, and classification as illustrated in Fig. \ref{fig:processing-flow}.\footnote{Github code will be made available upon publication}

\subsection{Dataset}

We used a publicly available dataset from the research paper "What Makes a Good Counselor? Learning to Distinguish between high-quality and Low-quality Counseling Conversations" ~\cite{perez-rosas-etal-2019-makes}.
This dataset contains a wide range of conversational topics, from clients feeling nervous about going to the mall with friends to discussing the desire to quit smoking. The dataset was compiled by identifying and analyzing video clips of counseling sessions sourced from platforms such as YouTube and Vimeo, focusing on those conducted using Motivational Interviewing. This study did not require a Research Ethics Board (REB) as the dataset is made available for non-commercial research purposes. 

For this research paper, we started with a dataset consisting of 262 conversational transcripts. Excluding nine no longer accessible or insufficient recording materials, the resulting 253 counseling conversations include 150 high-quality (HI) and 103 low-quality (LO) counseling sessions, as listed in Table \ref{tab:dataset_summary}. 

\subsection{Feature Extraction}
Using a feature engineering approach, we addressed the conversational quality from four key analytical dimensions: conversation dynamics, semantic similarity analysis, sentiment analysis, and question detection. A summary of these features is listed in Table \ref{tab:features}. In detecting and counting the questions asked in a conversation, we employed a specialized yet simple encoder model to identify questions. Question detection, sentiment analysis, semantic similarity, and other custom features were specifically tailored for engagement quality classification. These engagement features were designed to capture the degree to which the client actively participates in the session, responding to the therapist, asking questions, and demonstrating emotional or cognitive involvement.

\subsubsection{Conversational Dynamic Features}
An embedding was created to generate and compile conversational metrics to describe the interaction between the therapist and the client and to convert them into features that the model can use. These tailored features capture various dynamics of client-therapist interactions, including Interaction Duration, Turn-Taking Ratio, and Word Count Analysis. Word Count Analysis primarily revolves around the number of words spoken by the client or therapist in each utterance (turn), defined as \textit{Words Spoken per Turn}. When looking to evaluate the quality of client-therapist interactions, the length of turns taken by both speakers is extremely important as it gives insight into their conversational dynamics. A turn is defined as an uninterrupted sequence of words spoken by a speaker; using turns and the number of words spoken in a turn yields words spoken per turn. Words spoken per turn as a feature can be substituted for the average turn time of a speaker. By looking at turn lengths, the relative length difference between the client and the therapist can indicate more engagement from the client, which can mean the session is of high quality due to high participant engagement.

Utilizing the length of time someone spends speaking in combination with features like question/response bolsters the model in producing accurate results. The data for this feature was extracted using the YouTube transcript-api where the YouTube auto-generated transcripts from all original videos were extracted. Then the average duration for each speaking turn for these transcripts was recorded~\cite{perez-rosas-etal-2019-makes}. To maximize the data extracted from these transcripts further, standard deviation, skewness, and kurtosis are extracted from the words per turn metric by storing individual word counts of each turn in a list for this calculation. All these features put together and used by the model can give markers that are correlated with high-quality client-therapist interactions.

\subsubsection{Semantic Similarity Features}
The sentence similarity features used for this application are designed to analyze and score the similarities between the spoken words of the Therapist to those of the Patient. These features acted as indirect measures to analyze the topic alignment between the two speakers and helped verify a good conversation through relevancy. These features leverage the use of three pre-trained sentence embedding models that were selected based on their high performance in many benchmarks pertaining to sentence similarity. The three models used are: \textit{PromCSE}~\cite{jiang2022improved} -- a RoBERTa-based model which ranked as the top performing model in semantic textual similarity (STS) on the CxC and SICK~\cite{marelli-etal-2014-semeval} benchmarks; \textit{Sentence-BERT}~\cite{reimers2019sentencebert} -- a modification of the pre-trained BERT network which has been trained on the SICK and STS datasets for sentence similarity detection; and \textit{SAKIL sentence similarity semantic search} -- a fine-tuned version of PromCSE that has been trained on the Kaggle dataset. These three models have high-performance results on sentence similarity benchmarks and open-source availability to be used in our application. PromCSE was known for its clinical relevance, as it outperformed Sentence-BERT on medical dialogue benchmarks. Since there are no openly available embedding models that are fine-tuned for use in therapeutic settings, this multi-model approach was the most appropriate to achieve our classification goal. Ultimately, this approach made feature extraction more computationally heavy but reduced the risk of bias from any one model used on its own.

The three sentence embedding models produced a set of independent similarity scores for sentence pairs in the given therapy transcript. These similarity scores were used as features in the classification. The implementation of this semantic feature component is split into two sub-features: overall semantic similarity analysis and a conversation turn-order-aware semantic analysis. 

In the overall semantic similarity analysis approach, all of the sentences were embedded with a semantic similarity transformer. Then each sentence’s embedded vector,  $x_i$ for $i \in [1,N]$ where $N$ is the number of sentences, was compared to every other sentence vector in the conversation using a cosine similarity, $Sim(x_i, x_j)$.
\begin{equation}
    Sim(x_i, x_j)_{i \neq j \in N} = \frac{x_i \cdot x_j}{|x_i||x_j|}
\end{equation}
This comparison takes two different vectors $\{x_i, x_j\}$ as input and returns a value between 0 and 1 representing the similarity of the two vectors, a value closer to 1 indicates high similarity. Since we were comparing each sentence to all other sentences, this technique resulted in a matrix of $Sim1 \in R^{N(N-1)/2}$, where N is the number of sentences in the conversation. Statistical measures (mean, standard deviation, skewness, and kurtosis) of $Sim1$ were used as feature scores for the conversation. These four statistical scores combine to provide a detailed numerical view of the distribution of the similarity results. This process is repeated once for each of the three selected embedding models with four statistical measures, producing 12 unique feature scores. The goal of this approach is to analyze the consistency of the topic(s) discussed throughout the entire therapy session and the similarity in the word choice used throughout the session.

The conversation turn-order-aware semantic analysis considered the order in which the conversation took place when conducting semantic comparisons. This technique analyzed the conversation flow using the same sentence embedding extracted previously. Cosine similarities were applied to a smaller set of adjacent sentence pairs. Only the sentence that followed immediately was used in the comparison. In this sequential pairing approach, sentence 1 was compared with sentence 2, sentence 2 was compared with sentence 3, and so on. In a conversation with N sentences, this technique created a vector of $Sim2 \in R^{N-1}$ measure. Similarly to the overall analysis, each embedding model produced a set of 12 features. These features captured the conversation flow of each therapy session and indicated normal conversational patterns such as balanced back-and-forth exchanges between both parties, coherent topic transitions, and social elements were part of the conversation.  

With the combined use of three different sentence embedding models and two different techniques used to measure conversation content, the set of semantic similarity features provided global and local views of the conversation between the client and the therapist.

\subsubsection{Sentiment Features}
When trying to analyze the overall session outcome of any client-therapist interaction, understanding the sentiment of the conversation can provide a lot of insight. Sentiment analysis is a type of NLP task that can provide some emotional context to what different speakers are saying, usually labelling speech as either positive, neutral or negative. For this study, the \textit{Twitter-roBerta-base for Sentiment Analysis} model~\cite{Loureiro2022}, a RoBert model trained on Twitter exchanges and posted to the Hugging Face model hosting service, was chosen. Sentiment analysis was executed on every speaker's turn in the transcript, and a weighted average sentiment score was calculated for each conversation. Each speaker's turn was graded with a value of either 1, 2 or 3, representing negative, neutral, or positive sentiment, respectively. These sentiment values were weighted with the certainty of the estimation. A certainty of a sentiment score of 50\% was given less weight on the overall sentiment of the conversation. A sentiment turn feature was also created that kept track of how many times in the conversation the sentiment between the therapist and client changed. The more changes in the sentiment indicated higher motivation between the client and the therapist.

\subsubsection{Question Detection Feature}
The number and frequency of questions during a conversation can indicate the engagement level of participants. Asking questions to a therapist indicates a level of interest in the conversation and the subject matter that would indicate a high-quality therapy session. A question embedding was developed to identify and track questions asked by both therapists and clients during each conversation turn. Syntactically, a question is typically composed of a question word and an auxiliary verb. Based on this, any sentence that contains this structure or has a question mark punctuation will be identified as a question by the question embedding and is counted towards the various question detection metrics. This approach was necessary due to the inconsistent use of punctuation and question marks in the transcripts. This approach involved the creation of a word bank of both auxiliary verbs and question words through the use of a corpus \footnote{\url{https://github.com/kartikn27/nlp-question-detection}}. Transcripts are separated into sentences, which are further broken down into bi-grams, where each bi-gram is compared to its respective word bank. If a matching pair is found, then that portion is identified and counted as a question asked by the respective speaker. Turn by turn, the number of questions asked by the speaker is recorded and stored for the calculations needed to create features such as client questions per turn, therapist questions per turn, and the ratio of questions asked per turn.

\subsection{Data Processing}
Once the features were extracted, we followed a standard data exploratory process with a versatile approach, including normalization of data points to ensure consistency across features and handling of missing values with imputation. Furthermore, we addressed outliers and corrected structural errors to improve data quality and facilitate the effective utilization of our machine learning model.

\subsubsection{Normalization} Some classifiers, such as SVM, can be dominated by large-value features. Normalization removes feature influence on the model by ensuring all feature dimensions have the same scale. Tree-based classifiers like Random Forest or CatBoost are not affected by dominating features. To reduce the influence of large-value features, min-max data normalization was used to scale all features to the range of [0,1].

\subsubsection{Missing Data Imputation}
The features extracted from this imbalanced dataset are inherently imbalanced with various data ranges and missing features due to not meeting the threshold required to generate statistical values. Missing values were addressed by imputing them with the mean values of the respective features. Imputing missing values with the mean ensures that the dataset remains representative of the overall distribution of the features while minimizing the impact of missing data on the model training process. 

\subsubsection{Outlier Removal}
Isolation forest algorithm~\cite{liu2008isolation, liu2012isolation} with 100 isolation trees (proper binary trees) was used to detect outliers. Each isolation tree divides the instances recursively by randomly selecting a variable and a split value. The isolation forest computes the anomaly score $s(x,n)$ of an instance $x$ out-of-$n$ instances as

\begin{equation}
    s(s,n) = 2^{-
    \frac{E(h(x}{c(n}
    }
\end{equation}

where $E(h(x))$ is the average of path length $h(x)$ from a collection of isolation trees and $c(n)$ is the average depth of all $n$ instances. An anomaly is defined as when $E(h(x)) \rightarrow  c(n)$ resulting in  $ s \rightarrow 0.5$ border. On the other hand, if $x$ is isolated much earlier, i.e.,  its average path depth is much shallower than other instances $E(h(x)) << c(n)$, then the anomaly score $s(x,n) \rightarrow 1$. Or, if the average depth of $x$ is much deeper before isolation, $E(h(x)) >> c(n)$, then the anomaly score $s(x,n) \rightarrow 0$.

After removing 13 outliers (Table \ref{tab:dataset_summary}), the total number was reduced to 240, with 145 HI and 95 LO sessions. The dataset was then split into training and holdout test sets for cross-validation and model evaluation. To ensure balanced representation in the test set, equal numbers of samples (9 from each class) were randomly selected from both the HI (majority) and LO (minority) classes. High-quality conversations are defined as interactions between therapist and client that demonstrate meaningful dialogue, balanced participation, responsiveness, and effective therapeutic techniques~\cite{perez-rosas-etal-2019-makes}. The test sample size ($9x2=18$ samples) was calculated based on 10\% of the minority class size. Random samples were chosen from both classes to fulfill the required number of samples for the test set.

This approach was crucial to ensure that the trained model was evaluated on a representative test set while maintaining a balanced distribution of classes in the test data, thereby enhancing the reliability and fairness of the model evaluation process. Table \ref{tab:dataset_summary} summarizes the number of high- and low-quality counseling videos.

\begin{table}
    \caption{Summary of the dataset used in the study.}
    \label{tab:dataset_summary}
\centering
\resizebox{\linewidth}{!}{%
\begin{tabular}{lccc}
\hline
    &  \textbf{HI class} & \textbf{LO class} & \textbf{Total} \\
\hline
\\
   Passed Quality Assurance  & 150 & 103 & 253\\
   Outlier removed & 145 & 95 & 240 \\
   Holdout test set & 9 & 9  & 18\\
   Training set & 136 & 86 & 222 \\
\\
\hline
\multicolumn{4}{l}{\footnotesize{HI=high-quality. LO=low-quality.}} \\
\hline
    \end{tabular}}
\end{table}

\subsubsection{Data Augmentation}
After establishing the test set, the training set exhibited class imbalance, with 136 transcripts labelled as high-quality and 86 labelled as low-quality. Data resampling was applied to address the imbalanced data by downsampling the majority class. Data augmentation was used to project the performance of these features by upsampling the minority class to match the majority class samples as well as upsampling both classes (Table \ref{tab:smote-tomek-results}. We use a combination of Synthetic Minority Over-sampling Technique (SMOTE)~\cite{Chawla2002} and Tomek link(SMOTE+Tomek)~\cite{Batista2004} which uses SMOTE's linear interpolation to augment the minority class, and Tomek links' borderline nearest neighbour method to remove majority class instances that are closest to the minority samples. By combining SMOTE oversampling and Tomek links' undersampling, the dataset's class imbalance was effectively addressed. For analytical purposes, two additional upsamplings were used to augment the dataset to 200 samples and 300 samples in each class. By upsampling the data and creating additional datasets, we ensured that our classifier could be trained and evaluated on balanced datasets to improve performance, promote generalization capabilities, and remove bias toward specific classes. To illustrate that the original feature density was maintained, a set of kernel density estimation (KDE) plots can be found in Fig.~\ref{fig:KDE}. 

\begin{table}[htb]
\caption{SMOTE-Tomek augmented balanced datasets.}
\label{tab:smote-tomek-results}
\centering
\resizebox{\linewidth}{!}{%
\begin{tabular}{l c c}
\hline
\textbf{Dataset} & \textbf{HI Class} & \textbf{LO Class} \\
\hline
\\
Original Set & 145 & 95 \\
Balanced Holdout Test Set & 9 & 9 \\
Majority downsampled Training set & 86 & 86 \\
Minority upsampled datset 1 & 136 & 136 \\
Upsampled dataset 2 & 200 & 200 \\
Upsampled dataset 3 & 300 & 300 \\
\\
\hline
\multicolumn{3}{l}{\footnotesize{HI=high-quality. LO=low-quality.}}\\
\hline
\end{tabular}}
\end{table}

\subsection{Classification}
The classifier serves as a vital tool in quantifying the dynamic interplay between clients and therapists, shedding light on the effectiveness of the counseling sessions. We selected three classifiers that support hyperparameter tuning (Random Forest, CatBoost, and Support Vector Machine with a kernel) to validate these NLP features. These three classifiers cover a wide range of algorithms, where RF is based on bagging, SVM is based on hyperplane separation, and CatBoost for boosting methods. All three classifiers provided a suitable ground for hyperparameter tuning to enhance predictive accuracy, ensuring efficacy across a diverse array of engagement scenarios. 

Random Forest (RF) and CatBoost are tree-based ensemble classifiers that inherit the Decision Tree's versatility for both classification and regression tasks, dividing the feature space into regions and making predictions based on the majority class or average value within each region. RF constructs multiple decision trees during training and aggregates their predictions to improve robustness and accuracy. RF was chosen for its ability to handle a mixture of numerical and categorical features, scalability, robustness to overfitting, and ability to handle high-dimensional data. The integration of majority voting within the RF algorithm acts as a robust safeguard against uncertainties of data variability. By consolidating predictions from an ensemble of decision trees, the classifier created with Random Forest enhances overall accuracy and reliability, ensuring consistent performance even in the presence of outliers and addressing the intricate challenges inherent in assessing client-therapist interactions. 

CatBoost is an ensemble classifier that belongs to the Gradient Boosted Decision Trees family and thus also is known for its performance on heterogeneous data~\cite{hancock2020catboost}. The models are not affected by feature scaling and thus do require normalization. 

Support Vector Machine (SVM) is a powerful model used for classification and regression tasks, aiming to find the hyperplane that best separates classes while maximizing the margin between them. Unlike decision tree-based classifiers, data normalization is required when using SVM. With the usage of a kernel SVM classifiers are effective in high-dimensional spaces and suitable for linearly non-separable data. Four types of kernels were experimented within the SVM classifier: linear, Gaussian radial basis function (RBF), polynomial, and sigmoid. A set of hyperparameters, summarized in Table \ref{tab:hyperparameters}, was used to determine the best model using grid search for hyperparameter tuning. 

\begin{table}
    \caption{Classifier hyperparameters}
    \label{tab:hyperparameters}
    \centering
    \resizebox{\linewidth}{!}{%
    \begin{tabular}{ll}

    \multicolumn{2}{l}{\textbf{Random Forest}} \\
        n\_estimators & [50, 100, 200, 500] \\
        max\_depth &  [10, 15, 20, 30] \\
        min\_samples\_split &  [ [2, 5, 10, 20] \\
        min\_samples\_leaf &  [1, 2, 4, 8] \\
        \\
        \multicolumn{2}{l}{\textbf{CatBoost}} \\
        iterations  & [200, 500, 750] \\
        depth & [4, 6, 8], \\
        learning\_rate & [0.01, 0.05, 0.1] \\
        \\
        \multicolumn{2}{l}{\textbf{SVM}} \\
        C & [0.01, 0.1, 1, 10, 100] \\
        kernel  & ['linear', 'rbf', 'poly', 'sigmoid'] \\
        gammA  & [0.01, 0.1, 1, 'scale', 'auto'] \\

    \end{tabular}}
\end{table}

The cross-validation and test were performed sequentially in all datasets, i.e., the original down-sampled train dataset, the minority up-sampled datasets and the two classes up-sampled datasets. The Stratified K-Fold cross-validation technique with 5-fold ensured that each fold maintained the class distribution of the target variable. This mitigated the risk of overfitting and provided more reliable evaluation metrics. Hyperparameter tuning was applied on each fold to obtain 5 tuned models for each classifier. These models were then used on the holdout test set for performance measures.

\subsection{Performance Measure}
Trained classifiers were tested using the holdout test set and evaluated on the standard performance metrics for balanced data, including accuracy, precision, recall, F1--score, and AUC--ROC (Area Under the Receiver Operating Characteristic curve). To better understand the contribution of each feature, SHAP value ranking was applied to each classifier. SHAP analysis is an established technique for interpreting machine learning models and was employed to scrutinize individual predictions. The significance of each feature was measured by its SHAP value. Positive values indicate features that amplify the model's prediction, and negative values signify features that diminish it. Knowing SHAP's limitation~\cite{kumar2020problems}, the features were ranked using the mean SHAP value of all data with three hyperparameter-tuned classifiers. This approach provided us with a robust means of comprehending the influence of each feature on different models' performance. Mean SHAP values from the 5 tuned models were used to rank the feature importance. 

\section{Results}
\subsection{Data Preprocessing}
\label{sec:feature_analysis}
The total number of client turns within the transcripts was 5,073. Client turns represent instances where the client actively participates in the conversation or interaction, contributing to the overall dialogue. Similarly, the total number of therapists' turns was 5,138. Therapist turns denote instances where the therapist or counselor engages in the conversation, providing guidance, support, or feedback to the client.

A data frame was created to store features extracted from the original dataset. Missing values were addressed by imputing them with the mean values of the respective features. This approach was utilized because data was missing at random without any fixed pattern, and the missing values were relatively small (106 entries) in number compared to the overall dataset (11,132 entries). There was a small number of features (less than 1\%) that could not be extracted and thus were imputed with mean values. 

\begin{figure}[htb]
    \centering
    \includegraphics[width=\linewidth, trim={10 40 10 50}, clip]{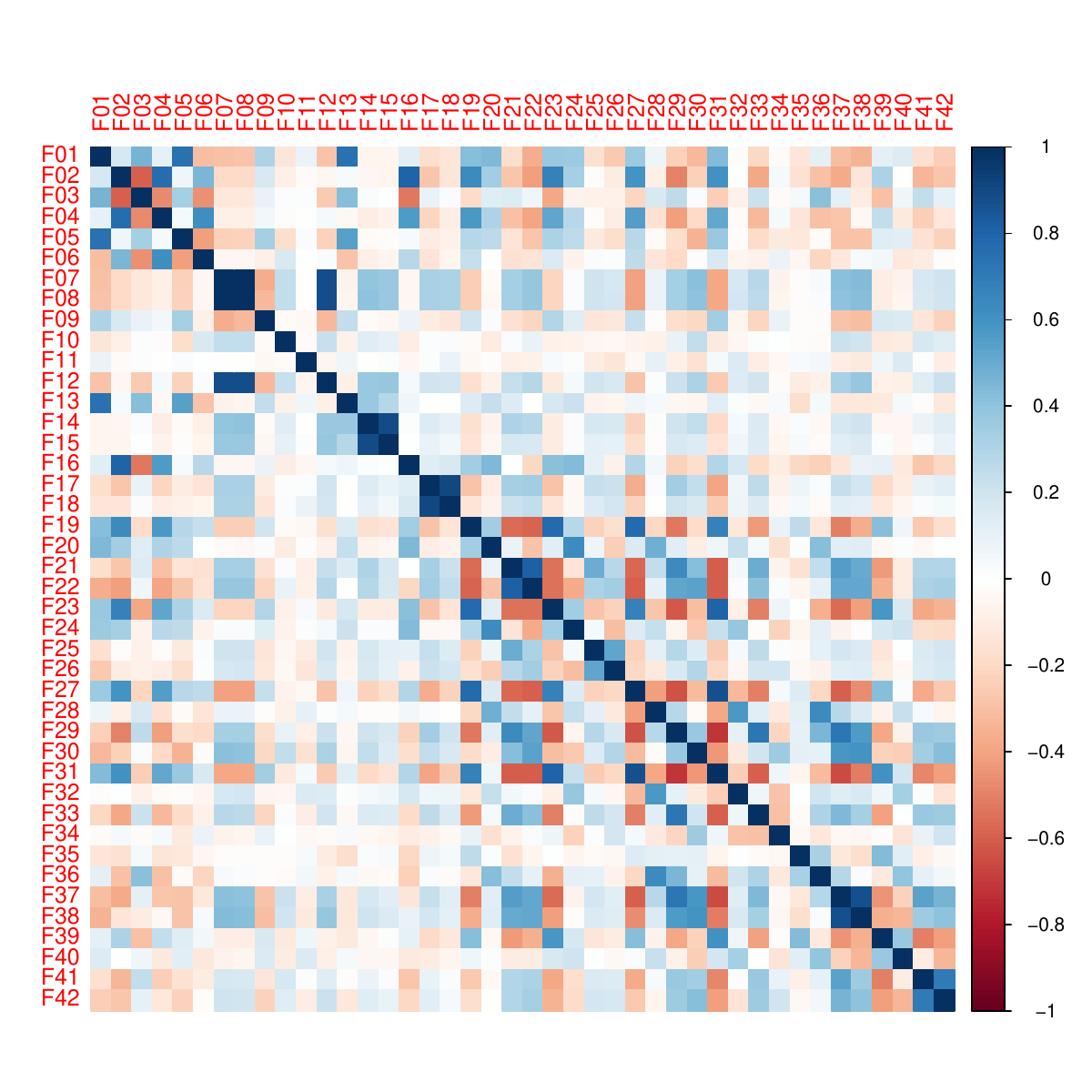}
        \caption{Feature correlation heatmap with highest value of 1 and lowest value of -0.71.}
        \label{fig:heatmap}
\end{figure}

As SHAP cannot address the correlation between features, Pearson correlation analysis revealed relationship patterns among features. The heatmap illustrated in Fig. \ref{fig:heatmap} shows the Pearson correlation between feature pairs with a minimum value of $r=-0.71$ indicating the strongest negative correlation. Fifteen of the correlation pairs were found to have strong positive correlations, as listed in Table \ref{tab:highcorrpairs}. The correlation coefficients between these pairs ranged between 0.75 to 1.0, indicating substantial inter-dependencies between certain feature pairs.

\begin{table}[htb]
   \caption{Fifteen strongest correlation feature pairs with Pearson correlation $r>0.75$}
    \label{tab:highcorrpairs}
    \centering
    \resizebox{\linewidth}{!}{%
    \begin{tabular}{|c|c|c|c|c|c|c|c|c|c|c|c|c|c|c|}
\hline
F07 & F17	& F14	& F07	& F08	& F37	& F27	& F21	& F23	& F02	& F19	& F19	& F02		\\
F08	& F18   & F15   & F12   & F12   & F38   & F31   & F22   & F31   & F16   & F27   & F23   & F04 \\
0.998   & 0.90      & 0.89      & 0.88  & 0.88  & 0.88  & 0.87  & 0.82  & 0.81  & 0.80  & 0.78  & 0.78  & 0.77 \\
\hline
    \end{tabular}}
\end{table}

\begin{table*}[htb]
\caption{Comparison of Classifier Performance on Holdout Test Set for all folds}
\label{tab:classification_results}
\centering
\resizebox{\linewidth}{!}{%
\begin{tabular}{lrrrrrl}
\hline
\textbf{Classifier} & \textbf{Acc} & \textbf{Pre} & \textbf{Rec} & \textbf{F1--Score} & \textbf{AUC}  & \textbf{Best Paraameters}\\
\hline
\\
\multicolumn{7}{l}{Majority downsampled dataset 0 (86)}\\
RF & 
76.7$\pm$6.1	& 73.4$\pm$6.3	 & 84.4$\pm$6.1 &	78.4$\pm$5.2	& 84.4$\pm$4.7 & \{M\_depth: 10, leaf: 1, split: 10, n\_est: 500\} \\
CB & 
74.4$\pm$8.4	 & 71.1$\pm$7.2	& 82.2$\pm$9.9 & 76.2$\pm$8.2	& 81.7$\pm$6.7 & \{depth: 8, iterations: 750, lr: 0.01\} \\
SVM & 
74.4$\pm$6.3	& 71.3$\pm$5.9	& 82.2$\pm$6.1	& 76.3$\pm$5.7	& 85.4$\pm$4.8 & \{C: 1, gamma: 1, kernel: 'rbf'\} \\
\\
\multicolumn{7}{l}{Minority upsampled dataset 1 (136)}\\
RF & 
87.8$\pm$2.5	 & 81.5$\pm$0.8	 & 97.8$\pm$5.0	 & 88.8$\pm$2.6	 & 93.1$\pm$3.1 &  \{M\_depth: 10, leaf: 1, split: 2, n\_est: 500\} \\
CB & 
81.1$\pm$6.3	 & 75.5$\pm$7.0	 & 93.3$\pm$6.1	 & 83.3$\pm$5.0	 & 89.1$\pm$2.7 & \{depth: 6, iterations: 750, lr: 0.1\} \\
SVM & 
76.7$\pm$6.1	 & 71.5$\pm$7.8	 & 91.1$\pm$5.0	 & 79.8$\pm$4.1	 & 87.9$\pm$3.2 & \{C: 10, gamma: 1, kernel: 'rbf'\} \\
\\
\multicolumn{7}{l}{Upsampled dataset 2 (200)}\\
RF & 
83.3$\pm$5.6	 & 77.1$\pm$5.8	 & 95.6$\pm$6.1	 & 85.2$\pm$4.6	 & 94.7$\pm$3.9 & \{M\_depth: 10, leaf: 1, split: 2, n\_est: 200\} \\
CB & 
84.4$\pm$4.7	 & 77.3$\pm$4.3	 & 97.8$\pm$5.0	 & 86.3$\pm$4.1	 & 92.8$\pm$4.6 & \{depth: 4, iterations: 750, lr: 0.1\} \\
SVM & 
78.9$\pm$6.1	 & 72.6$\pm$5.7	 & 93.3$\pm$6.1	 & 81.6$\pm$5.1	 & 92.8$\pm$2.8 & \{C: 100, gamma: 1, kernel: 'rbf'\} \\
\\
\multicolumn{7}{l}{Upsampled dataset 3 (300)}\\
RF & 
88.9$\pm$0.0	 & 81.8$\pm$0.0	 & 100.0$\pm$0.0	 
& 90.0$\pm$0.0 & 94.6$\pm$1.9 & \{M\_depth: 10, leaf: 1, split: 2, n\_est: 500\} \\
CB & 
83.3$\pm$5.6	 & 75.4$\pm$6.3	 & 100.0$\pm$0.0	 & 85.9$\pm$4.1	 & 91.9$\pm$2.8 & \{depth: 6, iterations: 500, lr: 0.1\} \\
SVM & 
81.1$\pm$5.0	 & 75.0$\pm$3.9	 & 93.3$\pm$6.1	 & 83.1$\pm$4.6	 & 93.6$\pm$1.0 & \{C: 10, gamma: 1, kernel: 'rbf'\} \\
\\
\hline
\multicolumn{7}{l}{\small{RF=Random Forest. CB=CatBoost}}\\
\multicolumn{7}{l}{\small{Acc=Accuracy. Pre=Precision. Rec=Recall. AUC=Receiver-operating characteristic curve Area under the curve.}} \\
\multicolumn{7}{l}{\small{n\_est=n\_estimators. M\_depth=max\_depth. leaf=min\_samples\_leaf. split=min\_samples\_split. lr=learning\_rate.}}\\

\hline
\end{tabular}
}
\end{table*}

\begin{figure*}[htb]
\centering
    \includegraphics[width=\linewidth]{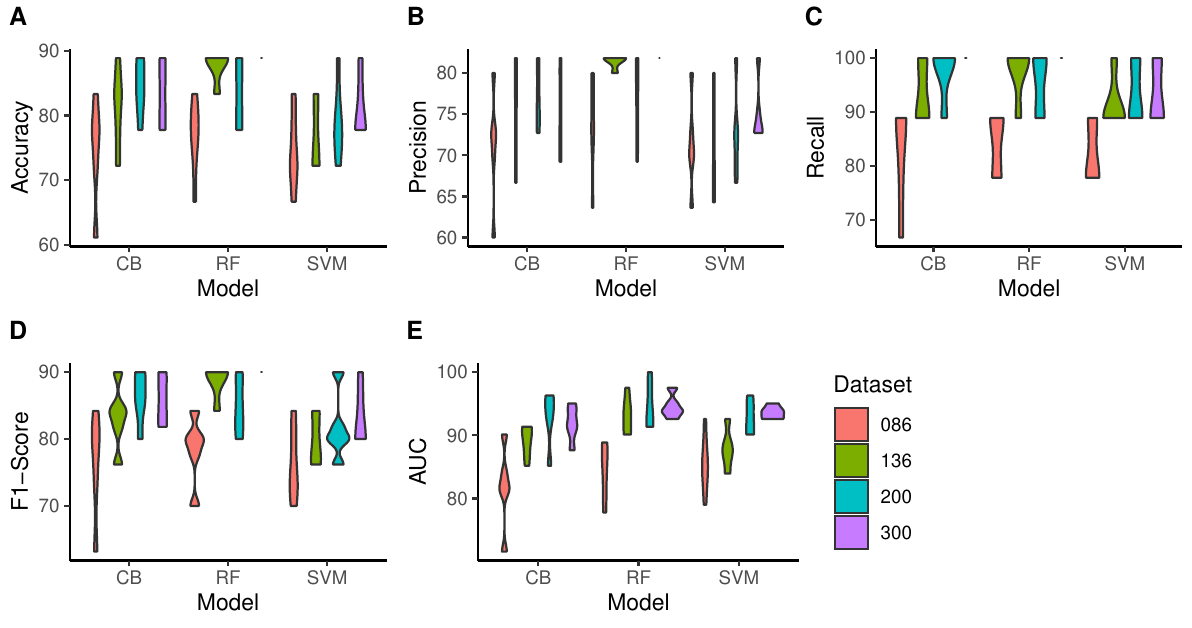}
    \caption{Performance measure illustrating the holdout test set on the cross-validation folds for Random Forest (RF), CatBoost (CB) and Support Vector Machine (SVM). The holdout test set has only 18 samples, with 9 from each class. Instead of providing the best result, the test set was used in each cross-validation fold after hyperparameter tuning.}
    \label{fig:metrics_comparison}
\end{figure*}

\subsection{Classifier Performance}
The classification results using hyperparameter-tuned classifiers are presented in Table~\ref{tab:classification_results} and illustrated in Fig.~\ref{fig:metrics_comparison} for the balanced datasets. As the holdout test set only contained 18 data points, we applied the test set in each cross-validation fold for comparison. The holdout test performance on the 5-fold with hyperparameter-tuned models is summarized in Table \ref{tab:classification_results}. Using AUC Score as the overarching measure, RF and SVM alternately outperformed each other depending on the dataset. SVM performed best on dataset 3 and showed strong AUC performance on dataset 3 (93.6\%), while RF showed the most consistent gains in both accuracy and F1-score across datasets 0, 1 and 3.

Fig.~\ref{fig:metrics_comparison} presents a comprehensive comparison of classifier performance across different dataset variants. The analysis reveals several key findings:

\subsubsection{Overall Performance Trajectory}
All classifiers benefited from the SMOTE-Tomek augmentation, with performance improving as the dataset size increased. The best performance was observed in the 300-sample upsampled dataset (dataset 3), where Random Forest achieved the highest accuracy (88.9\%), F1 score (90.0\%), and AUC (94.6\%). CatBoost followed closely with an F1 score of 85.9\% and 100\% recall.
   
\subsubsection{Classifier-Specific Performance}
\textbf{SVM} demonstrated balanced, but slightly lower, performance across all metrics, achieving up to 81.1\% accuracy, 75.0\% precision, 93.3\% recall, and 93.6\% AUC in the largest upsampled dataset.\textbf{CatBoost} delivered stable performance with F1 scores up to 85.9\%. \textbf{Random Forest} showed the highest recall (100\%) in dataset 3, though with more variability in precision and F1.
   
\subsubsection{Metric-Specific Trends}
\textbf{AUC Scores} showed consistent improvement across dataset iterations, with the highest value of 94.7\% observed for Random Forest in the 200-sample dataset and 93.6\% for SVM in the 300-sample dataset. \textbf{Precision} values improved with augmentation, though they remained moderate overall, peaking at 81.8\% for Random Forest, 75.4\% for CatBoost, and 75.0\% for SVM. \textbf{Recall} increased more substantially, with both Random Forest and CatBoost achieving 100\% recall in the final dataset. \textbf{F1 Scores} showed steady growth across all classifiers, reaching maximums of 90.0\% for Random Forest, 85.9\% for CatBoost, and 83.1\% for SVM.

\subsection{Feature Contribution}
To maximize the usage of the available sample. The original unaugmented dataset, with 136 and 86 training samples in HI and LO classes, was used to determine the type of features contribute more. Table \ref{tab:SHAP} ranks the top contributing features and the corresponding feature type for each classifier. Overall, $17/30$ features belong to semantic analysis, $12/30$ features from the conversation analysis, and one from the question detection.

\begin{table}[htb]
    \centering
       \caption{SHAP value ranking}
    \label{tab:SHAP}
    \begin{tabular}{lcccccccc}
    \hline
\textbf{Rank}	& \textbf{RF}	&  \textbf{Type}	& & 
 \textbf{CB}	&  \textbf{Type}	& &  \textbf{SVM}	& \textbf{Type} \\
 \hline\\
1	& F16	& Conv	& & F16	& Conv	& & F16	& Conv \\
2	& F02	& Conv	& & F02	& Conv	& & F32	& Sem \\
3	& F07	& Conv	& & F07	& Conv	& & F12	& Conv \\
4	& F32	& Sem	& & F32	& Sem	& & F28	& Sem \\
5	& F12	& Conv	& & F28	& Sem	& & F16	& Conv \\
6	& F24	& Sem	& & F35	& Sem	& & F25	& Sem \\
7	& F28	& Sem	& & F30	& Sem	& & F37	& Sem \\
8	& F20	& Sem	& & F01	& Conv	& & F23	& Sem \\
9	& F04	& Ques	& & F22	& Conv	& & F36	& Sem \\
10	& F36	& Sem	& & F26	& Sem	& & F33	& Sem \\
\\
\hline
\multicolumn{9}{l}{\footnotesize{Conv=conversational dynamics. Sem=Semantic similarity.}}\\
\multicolumn{9}{l}{\footnotesize{Ques=Question detection.}}\\
\hline
    \end{tabular}
 
\end{table}

\section{Discussion}

The findings from this study underscore the utility of a multidimensional NLP framework for assessing engagement during therapeutic sessions. By leveraging advanced NLP techniques, such as sentiment analysis, semantic similarity, and conversational dynamics, the proposed framework introduces a novel approach to quantifying the quality of therapist-client interactions. Utilizing a dataset of 253 motivational interviewing session transcripts, the framework employed a Random Forest algorithm to classify engagement levels. SMOTE oversampling was applied to address the class imbalance, ensuring a more equitable representation of the unengaged class. This integration of linguistic and behavioral features into a comprehensive engagement score presents significant implications for enhancing both virtual and in-person mental health care.

This study also demonstrated the strong predictive power of semantic features in classifying engagement. SHAP analysis revealed that 17 of the top 30 most important features across all classifiers originated from semantic similarity metrics, with only one feature from question detection. These results suggest that semantic flow and topic coherence (the extent to which therapist and client stay aligned in their exchanges) are more critical to engagement prediction than superficial markers like question frequency.

One of the critical insights derived from this study is the role of conversational flow and emotional tone in engagement assessment. Prior studies have demonstrated the predictive power of sentiment and semantic analysis in fields such as education and social robotics~\cite{lee2018, Liu2022, Pattar2019}. Features like the turn-taking ratio showed a strong correlation with therapist-rated engagement (r = 0.65, p $<$ 0.01), validating their clinical utility. Our findings align with this body of work, emphasizing features like turn-taking ratios and semantic coherence as essential markers of engagement. These results deepen our understanding of therapeutic dynamics and lay the groundwork for developing real-time feedback systems to assist clinicians in optimizing treatment strategies.

Turn-taking is a conversational feature linked to engagement, particularly for therapy sessions. When clients have a more equal share of conversational turns, it demonstrates their active involvement. In contrast, highly imbalanced conversations, where the therapist overly controls the topics, may indicate poor engagement or client disengagement. In motivational interviewing, successful sessions often involve a structured but flexible exchange, where the therapist’s contributions guide the conversation without overpowering the client's voice. By incorporating turn-taking ratios, our framework aligns with these psychological principles, demonstrating a novel approach to quantifying engagement beyond subjective impressions.

However, an ongoing debate in computational research centers on whether machines can truly comprehend human emotions beyond surface-level linguistic cues~\cite{Robinson2009}. While the NLP-based framework effectively categorizes engagement levels, it lacks the contextual depth that human therapists provide. Therapists’ cognitive abilities allow them to interpret patient history, subtle cues, and emotions, often filtered through their lived experiences. In contrast, NLP systems could supplement therapy by incorporating multimodal data, such as video, audio, and movement, alongside patient history. Nevertheless, these tools are not intended to replace clinicians but rather to augment their capabilities, as supported by previous literature on digital mental health interventions.

The framework also has the potential to improve the efficiency and efficacy of therapy sessions by providing actionable insights to guide treatments. Engagement is a cornerstone of therapeutic practice, influencing how therapists adapt their approaches to individual patient needs~\cite{Gellatly2019}. Without adequate engagement, therapy risks devolving into a one-sided lecture that fails to address the patient’s multifaceted emotional landscape. For example, Harrison et al.~\cite{Harrison2019} found that patient-reported expected engagement predicted greater symptom improvement in cognitive-behavioral therapy (CBT), underscoring the importance of understanding and fostering engagement.

Despite these advancements, significant ethical concerns arise in the application of NLP-based methodologies to mental health care. Future deployments will use federated learning to ensure patient data remains on-premises, addressing privacy concerns in sensitive healthcare settings. Therapy involves sensitive and deeply personal information, necessitating stringent privacy protections. Patients must be fully informed about how their data is stored, processed, and shared by NLP systems. Furthermore, therapists must receive adequate training in digital tools to effectively communicate their use and limitations to patients. This transparency is critical in ensuring that patients feel comfortable and retain agency in their treatment process.

While the use of Random Forest ensembles bolstered the reliability of engagement predictions, certain limitations persist. The reliance on text-based transcripts overlooks non-verbal cues, such as facial expressions and voice intonations, that are integral to understanding engagement. A preliminary analysis of 20 video sessions using OpenFace 2.0 revealed that facial expressiveness (AU12, AU25) improved engagement prediction accuracy by 12\%~\cite{Baltrusaitis2018}. Future research should aim to incorporate multimodal data to achieve a more comprehensive assessment. Additionally, the dataset’s focus on motivational interviewing limits the generalizability of findings to other therapeutic modalities and cultural contexts. Expanding the dataset to include diverse therapy approaches and multilingual transcripts would enhance the framework’s applicability. Exploring advanced data augmentation techniques beyond SMOTE, such as generative models, could further improve model robustness and performance. Finally, while Twitter-RoBERTa is trained in social media language, therapy transcripts contain different linguistic patterns, such as longer, structured responses, professional tone, and mental health-specific terminology. This mismatch could introduce biases, particularly if informal speech patterns in therapy sessions are misclassified. Future work may be improved if models are trained on healthcare or therapy-specific texts to improve domain specificity.

\section{Conclusion}
This framework provides clinicians with actionable, real-time feedback on therapeutic engagement, filling a critical gap in virtual mental healthcare. By bridging NLP and clinical expertise, it advances toward AI-augmented, not replaced, therapeutic practice. The findings demonstrated that machine learning models, particularly Random Forest and CatBoost, can achieve strong performance, with accuracies up to 88.9\%, F1 scores up to 90.0\%, and AUCs as high as 94.6\%. To produce a classifier that can generate accurate labels that measure client engagement in a therapeutic setting, the following feature groups were selected: turn-taking ratio, verbal interaction, topic transition, client feedback, emotional tone, sentiment analysis, and semantic similarity. Features capturing semantic coherence and topic alignment were consistently the most predictive across all classifiers, suggesting that high-quality therapy sessions are more strongly marked by meaningful, on-topic exchanges than by surface-level metrics like question frequency.

This is an important topic to pursue since improving in this sector will allow clients and therapists to have more efficient and productive sessions. This increase in productivity and participation can foster a better therapeutic relationship and have a positive outcome in the field. As digital mental health continues to grow, tools like this framework can help maintain care quality, identify disengagement early, and support therapists in delivering more responsive, personalized treatment.

\appendix

\subsection{Data Augmentation}
\label{App:DataAugmentation}
\begin{figure*}[htb]
     \centering
        \includegraphics[width=\linewidth]{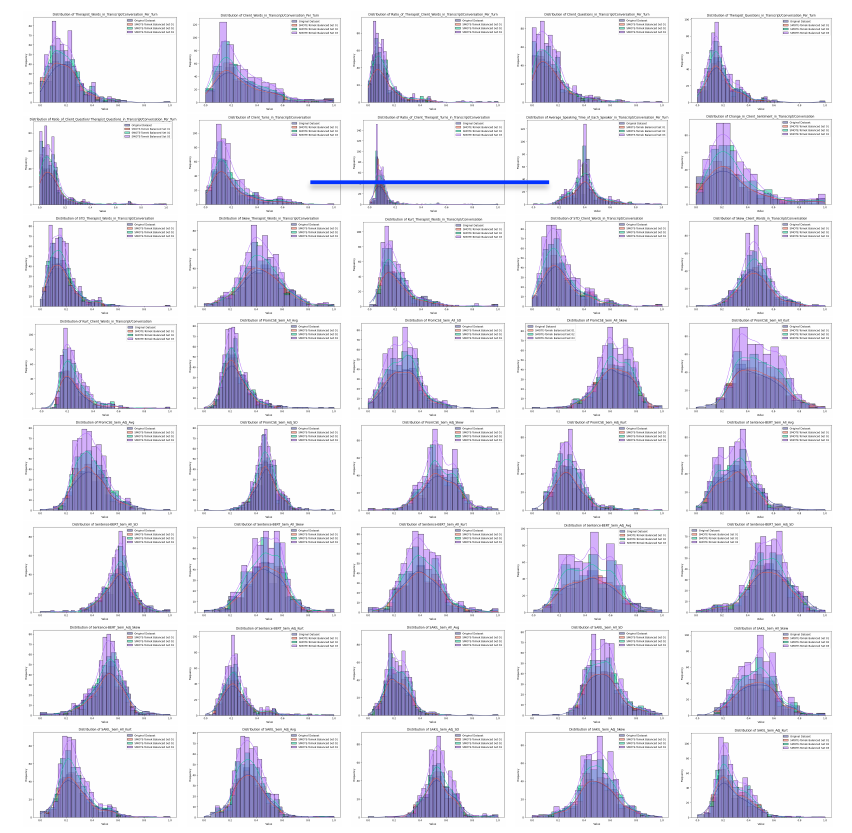}
    \caption{Kernel density estimation of the balanced features using SMOTE+Tomek.}
    \label{fig:KDE}
\end{figure*}

Handling imbalanced classification is important in machine learning when one class (the minority class) is significantly underrepresented compared to another class (the majority class). Imbalanced classification can lead to biased models that perform poorly in the minority class. Fig. \ref{fig:KDE} provides the kernel density plot of the features after augmentation results.

\newpage

\subsection{Feature Definition Table}

A list and descriptions of all the conversational dynamic features can be found in Table VII.

\begin{table*}[htb]
   \caption{Four feature dimensions}
    \label{tab:features}
    \centering
    \begin{tabular}{|p{6cm} |p{9.5cm} |l|}
    \hline
    \textbf{Feature} &  \textbf{Description} & \textbf{ID}\\
    \hline
    \multicolumn{3}{|l|}{\textbf{Conversational Dynamic Features}} \\
    \hline
    Therapist Words in Transcript/Conversation Per Turn  & Average therapist words uttered per turn in a transcript  & F01\\
        \hline
Client Words in Transcript/Conversation Per Turn & Average client words uttered per turn in a transcript  & F02\\
    \hline
Ratio of Therapist Client Words in Transcript/Conversation Per Turn & A ratio of client turns and therapist turns in a transcript  & F03\\
    \hline
Client Turns in Transcript/Conversation & Number of turns taken by a client in a transcript/conversation  & F07\\
    \hline
Therapist Turns in Transcript/Conversation & Number of turns taken by a therapist in a transcript/conversation &  F08\\
    \hline
Ratio of Client Therapist Turns in Transcript/Conversation & Number of turns taken by a therapist in a transcript/conversation & F09\\
    \hline
 Average Speaking Time of Each Speaker in Transcript/Conversation Per Turn & The average speaking time of each speaker turns. The time for each turn was taken from the auto generated YouTube transcripts. & F10\\
     \hline
STD/Skew/Kurt Therapist Words in Transcript/Conversation & The STD/Skew/Kurt between the number of words spoken in every utterance by the therapist in a transcript. & F13-F15\\
    \hline
STD/Skew/Kurt Client Words in Transcript/Conversation & The STD/Skew/Kurt between the number of words spoken in every utterance by the client in a transcript.& F16-F18 \\
    \hline
    
\multicolumn{3}{|l|}{\textbf{Semantic Similarity Features}} \\
\hline
    
MODEL A Sem All Avg/SD/Skew/Kurt & Using the SAKIL sentence transformer to find the mean/SD/Skew/Kurt value of the semantic similarity comparisons between each sentence to every other sentence in the conversation. & F35-F38\\
    \hline

MODEL A Sem Adj Avg/SD/Skew/Kurt & Using the SAKIL sentence transformer to find the mean/SD/Skew/Kurt value of the semantic similarity comparisons between each sentence to the following sentence in the conversation. & F39-F42\\
    \hline
MODEL B Sem All Avg/SD/Skew/Kurt & Using the PromCSE sentence transformer to find the mean/SD/Skew/Kurt value of the semantic similarity comparisons between each sentence to every other sentence in the conversation. & F19-F22\\
    \hline
MODEL B Sem Adj Avg/SD/Skew/Kurt & Using the PromCSE sentence transformer to find the mean/SD/Skew/Kurt value of the semantic similarity comparisons between each sentence to the following sentence in the conversation. & F23-F26\\
    \hline  
MODEL C Sem All Avg/SD/Skew/Kurt & Using the Sentence-BERT sentence transformer to find the mean/SD/Skew/Kurt value of the semantic similarity comparisons between each sentence to every other sentence in the conversation. & F27-F30\\
    \hline
MODEL C Sem Adj Avg/SD/Skew/Kurt & Using the Sentence-BERT sentence transformer to find the meanSD/Skew/Kurt value of the semantic similarity comparisons between each sentence to the following sentence in the conversation. & F31-F34\\
    \hline
 
 \multicolumn{3}{|l|}{\textbf{Sentiment Features}} \\
\hline
    
Overall Sentiment of Client & A feature that measures the sentiment of each speaker turns for
the client and returns a sentiment score of 3 for positive, 2 for neutral or 1 for negative. These scores are then averaged to get an overall sentiment score for the conversation. & F11 \\
    \hline
Change in Client Sentiment in Transcript/Conversation & An extension of sentiment analysis that finds the number of times the sentiment of the client changes between speaker turns & F12\\
    \hline
    
\multicolumn{3}{|l|}{\textbf{Question Detection Features}} \\
    \hline
    Client Questions in Transcript/Conversation Per Turn & The average number of questions asked by the client per response in a transcript  & F04\\
    \hline
Therapist Questions in Transcript/Conversation Per Turn  & The average number of questions asked by the therapist per response in a transcript & F05 \\
    \hline
Ratio of Client Question/Therapist Questions in Transcript/Conversation Per Turn & The ratio of client questions to therapist questions in a transcript  & F06\\
    \hline
    \end{tabular}
\end{table*}

\section*{Acknowledgment}
The authors would like to thank the Natural Sciences and Engineering Research Council of Canada (NSERC), New Frontiers in Research Fund, and LeaCros for the funding.

\bibliographystyle{IEEEtran} 
\bibliography{reference.bib}

%
%
%
\end{document}